\documentclass[sigconf]{acmart}

\AtBeginDocument{%
  \providecommand\BibTeX{{%
    \normalfont B\kern-0.5em{\scshape i\kern-0.25em b}\kern-0.8em\TeX}}}

\acmConference[KDD '23]{Make sure to enter the correct
  conference title from your rights confirmation emai}{August 06--10,
  2023}{Longbeach, CA}
\acmBooktitle{KDD '23, Workshop on Data Science for Social Good, 2023, Longbeach, CA}

\usepackage{times}
\usepackage{soul}
\usepackage{url}
\usepackage[utf8]{inputenc}
\usepackage{graphicx}
\usepackage{amsmath}
\usepackage{amsthm}
\usepackage{booktabs}
\usepackage{algorithm}
\usepackage{algorithmic}
\usepackage[switch]{lineno}
\usepackage{xcolor}
\usepackage{diagbox}
\usepackage{booktabs}

\begin{document}

\title[A Preliminary Investigation in Low-Listenership Prediction for Kilkari Mobile Health Program]{Analyzing and Predicting Low-Listenership Trends in a Large-Scale Mobile Health Program: A Preliminary Investigation}

\author{Arshika Lalan}
\affiliation{%
  \institution{Google Research}
  \country{India}
  }
\email{arshikal@google.com}

\author{Shresth Verma}
\affiliation{%
  \institution{Google Research}
  \country{India}
  }
\email{vermashresth@google.com}

\author{Kumar Madhu Sudan}
\affiliation{%
  \institution{ARMMAN}
  \country{India}
  }
\email{madhu@armman.org}

\author{Amrita Mahale}
\affiliation{%
  \institution{ARMMAN}
  \country{India}
  }
\email{amrita@armman.org}

\author{Aparna Hegde}
\affiliation{%
  \institution{ARMMAN}
  \country{India}
  }
\email{aparnahegde@armman.org}

\author{Milind Tambe}
\affiliation{%
  \institution{Google Research}
  \country{India}
  }
\email{milindtambe@google.com}

\author{Aparna Taneja}
\affiliation{%
  \institution{Google Research}
  \country{India}
  }
\email{aparnataneja@google.com}

\renewcommand{\shortauthors}{Lalan et al.}


\begin{abstract}
Mobile health programs are becoming an increasingly popular medium for dissemination of health information among beneficiaries in less privileged communities. Kilkari is one of the world's largest mobile health programs which delivers time sensitive audio-messages to pregnant women and new mothers. We have been collaborating with ARMMAN, a non-profit in India which operates the Kilkari program, to identify bottlenecks to improve the efficiency of the program. 
In particular, we provide an initial analysis of the trajectories of beneficiaries' interaction with the mHealth program and examine elements of the program that can be potentially enhanced to boost its success. We cluster the cohort into different buckets based on listenership so as to analyze listenership patterns for each group that could help boost program success .
We also demonstrate preliminary results on using historical data in a time-series prediction to identify beneficiary dropouts and enable NGOs in devising timely interventions to strengthen beneficiary retention.

\end{abstract}

\maketitle

\section{Introduction}
Mobile Health (mHealth) programs make use of mobile phone devices to deliver healthcare services while raising awareness about critical information and knowledge that contributes to optimal health outcomes. They play a key role in making healthcare more accessible for the less privileged \cite{mccool2022mobile,wood2019taking,guo2020mobile}.

In this paper, we focus on our collaboration with ARMMAN \cite{armman-mhealth}, an India-based non-profit organization which conducts mHealth programs in India to increase awareness during antenatal and postpartum care for pregnant women and mothers. It has adopted mHealth interventions to assist in bringing down maternal and child mortality rates. mMitra, an initiative by ARMMAN, is a free mobile call service that delivers organized preventive care information to enrolled women throughout their pregnancy and child infancy, on a weekly/bi-weekly basis. mMitra has successfully deployed SAHELI \cite{51839}, a system to efficiently utilize the limited capacity of healthcare resources to boost engagement with the program.

In 2016, the Ministry of Health \& Family Welfare (MoHFW) launched the Kilkari program, a free mobile health (mHealth) education service that sends women preventive care information during pregnancy and child infancy. Kilkari is an IVR service designed to deliver weekly pre-recorded, stage-specific audio messages to pregnant women and mothers with children under the age of 1 year. Currently operational in 18 states and Union Terriroties, Kilkari has reached over 30 million women and their children to date, and has 3 million active subscribers. ARMMAN is a technical, content \& creative production, and implementation partner to the MoHFW in making Kilkari available pan-India.

Like most mHealth programs \cite{madanian2019mhealth,aranda2014systematic}, Kilkari continues to evolve with improvements in technology and infrastructure. %
We discuss some unexplored questions based on these new developments in Kilkari. %
Our key contributions are : 
\begin{itemize}
    \item \textbf{Analyzing listenership patterns of different beneficiary buckets:} 
    We segment our cohort of beneficiaries into four buckets based on listenership (High Pickup Rate - High Engagement Rate, High Pickup Rate - Low Engagement Rate, Low Pickup Rate - High Engagement Rate, Low Pickup Rate - Low Engagement Rate ) and analyse each bucket individually. This segmentation helps us analyze the problem of listenership using two metrics - pickup and engagement - which we show are equally important. While previous works have only used pickup rates as an indication of listenership, we show that high pickup rates could be coupled with low engagement rates and thus should be taken into account to bolster program success.
    \item \textbf{Analyzing the impact of time slot on call pickup and call engagement:} 
    For our analysis, we construct seven time slots in the interval from 8 AM to 10 PM, divided into two-hour intervals. All calls, including reattempts, are put into one of these time slots. Our findings show that the beneficiaries in different buckets tend to exhibit a preference for some time slots over the others, which can be useful in boosting listenership
    \item \textbf{Predicting low-listeners:} 
   Our final contribution is predicting engagement rates (the rate at which beneficiary listens to a call for more than 30 seconds) along with pickup rates (the rate at which beneficiary picks up calls) of the entire cohort using
    preliminary time-series modelling. We showcase how to make such predictions using only listenership trajectories and no demographic features, which can be useful in programs with sensitive and limited beneficiary information, a key characteristic of most mHealth programs. Additionally, predicting low-listenership of beneficiaries from historical data can assist NGOs in planning timely interventions to improve beneficiary retention.
\end{itemize}
 Through the study of Kilkari as a real-world use case, our analysis offers insights that can be extended to other large-scale mHealth programs.

\section{Related Works And Background}
Adherence monitoring in healthcare is an extensively studied problem for diseases like HIV \cite{HIV}, cardiac problems \cite{son2010application}, Tuberculosis \cite{Killian_2019,10.1001/archinte.1996.00440020063008}, etc.
In mobile health programs as well, techniques such as sequential modelling \cite{nishtala2021selective} and markov decision process models \cite{mate2022field,Killian_2019} have been used to improve beneficiaries adherence to audio messages.

Kilkari is the largest maternal mHealth messaging program in the world \cite{KilkariArticle}. Given the scale of Kilkari's outreach, it has been a subject of multiple research studies in the past \cite{mohan2022optimising,chamberlain2021ten}. Infrastructural and technological investments have led to the successful resolution of some of the key challenges discussed in previous works, while others have become less significant in the course of time.  

\cite{chamberlain2021ten} address the process of scaling Kilkari successfully across geographies. At scale, Kilkari was redesigned to make calls throughout the day as opposed to based on subscriber preference to increase cost-efficiency. An unintentional consequence was that sometimes calls were made at times that could prove unsuitable to beneficiaries. %
~\cite{chamberlain2021ten} also highlight certain considerations that need to be taken into account in achieving success at scale among the rural population, such as sim churn and gender gap in mobile phone access and digital literacy. ~\cite{mohan2021can} investigate the impact of variations in phone ownership levels among different social strata, such as income and education. %
The paper highlights that a successfully picked call may not guarantee that it was answered by the intended recipient, as it may have been answered by another unaware family member. Consequently, it would be beneficial to analyse call engagement (duration of call listened to) in conjunction with pickup rate to measure the outcome of the program, as opposed to analysing pickup rates exclusively. 
Our work represents the first attempt in predicting both pickup rates and engagement rates for program beneficiaries. %

The key contribution of this paper centers around the 
challenges that Kilkari faces presently, in particular emphasizing predicting and contacting beneficiaries in their preferred time slots, as well as prioritizing program engagement rates along with pickup rates. Finally, we use predictive time-series modelling to predict engagement rates and pickup rates of the cohort. \cite{shahpreliminary} also uses a predictive model for low listenership prediction. However, our work uses both engagement and call pickup as metrics for low listenership, which our secondary analysis from bucketing beneficiaries shows are equally important metrics. We thus learn separate binary classification models for both these target variables. Moreover, our secondary analysis also underscores factors such as time slot of placing calls and technical success rate of calls which can be important for boosting pickup and engagement rates for the low listener groups, which has not been highlighted by \cite{shahpreliminary}. 

\section{Problem Formulation}
\subsection{Data Description}
The Kilkari mobile health program is an outbound service that makes periodic automated voice calls to pregnant women and new mothers%
, starting from the second trimester of pregnancy until the time the child is one year old. One voice message is scheduled for every beneficiary in each of these 72 weeks, which contains information on topics such as maternal and child health, immunization and family planning. 

Each beneficiary receives a maximum of 9 call attempts over 4 days, until they pick up the phone to increase the probability of them picking up the call. For every call attempt, the time and day of the attempt as well as the technical status is noted in the call records database. The technical status gives information on whether the call was picked up or the line was busy, switched off or out of network coverage.

In our analysis, we consider beneficiaries enrolled in the Kilkari program in the Indian state of Orissa between the year 2020-2022. Particularly, we take the set of 240K beneficiaries registered in Orissa and who received a call in the first week of 2022. We then create a dataset of their call history trajectory containing information on call attempts number, date and time of call, gestational age of beneficiary at the time of call, technical status of call and duration of the call.

\subsection{Time Slot Analysis}
Every week, a beneficiary receives the first call attempt on the same day as the day of their Last Menstrual Period date. If the beneficiary doesn't pick up the call, the call is attempted again.%

However, %
the time of receiving a call in a day is currently randomized. In our analysis, we demonstrate the value in predicting and utilizing a favourable time slot for beneficiaries.

\subsection{Low-Listenership Prediction}
Low-listenership of beneficiaries can be characterised using various metrics. %
Based on discussions with domain experts, we consider two definitions of low-listenership:
\begin{enumerate}
    \item low-pickup rate: beneficiaries who have picked up less than 3 calls within a 6 week time window
    \item low-engagement rate: beneficiaries who have engaged with a call less than 3 times within a 6 week time window. Engagement is defined as listening to a call for more than 30 seconds (average message length is typically around 90 seconds).
\end{enumerate}

Based on these definitions, we consider the following problem: Can a prediction be made in advance on when a beneficiary will become a low-listener. In such a case, early intervention by the NGO can help keep beneficiaries engaged with the health-information calls in the long run.

This low-listenership prediction problem can be formulated as a time-series prediction task. Starting at some point in time, we use history of beneficiaries' listenership trajectory for $N_{features}$ weeks to predict low-listenership $N_{offset}$ weeks in the future. To convert the raw dataset described previously for the time series prediction task, we split listenership trajectory of all beneficiaries into multiple rolling windows. Specifically, each window is of length $N_{features}+N_{offset}+6$ weeks where the last 6 weeks are used for defining the low-listenership binary flag.
For each beneficiary, we create time series of duration of calls every week, number of attempts, status of calls, and the date and time of calls. Given these temporal features, we try to predict whether a beneficiary would have low-listenership. Note that beneficiaries enter and exit the Kilkari program at different points in time. Thus, we would get different number of time-windows for different beneficiaries.

Finally, we split the beneficiaries into train ($80\%$) and test ($20\%$) sets, and thus create the train and test time series datasets. Splitting on beneficiaries rather than temporally allows us to learn a prediction model that can predict low-listenership for new beneficiaries which enter the system and have at least $N_{features}$ weeks of call history.

\section{Experiments}
\subsection{Measuring efficacy of each successive attempt}
\begin{figure}[h!]
  \includegraphics[width=\linewidth]{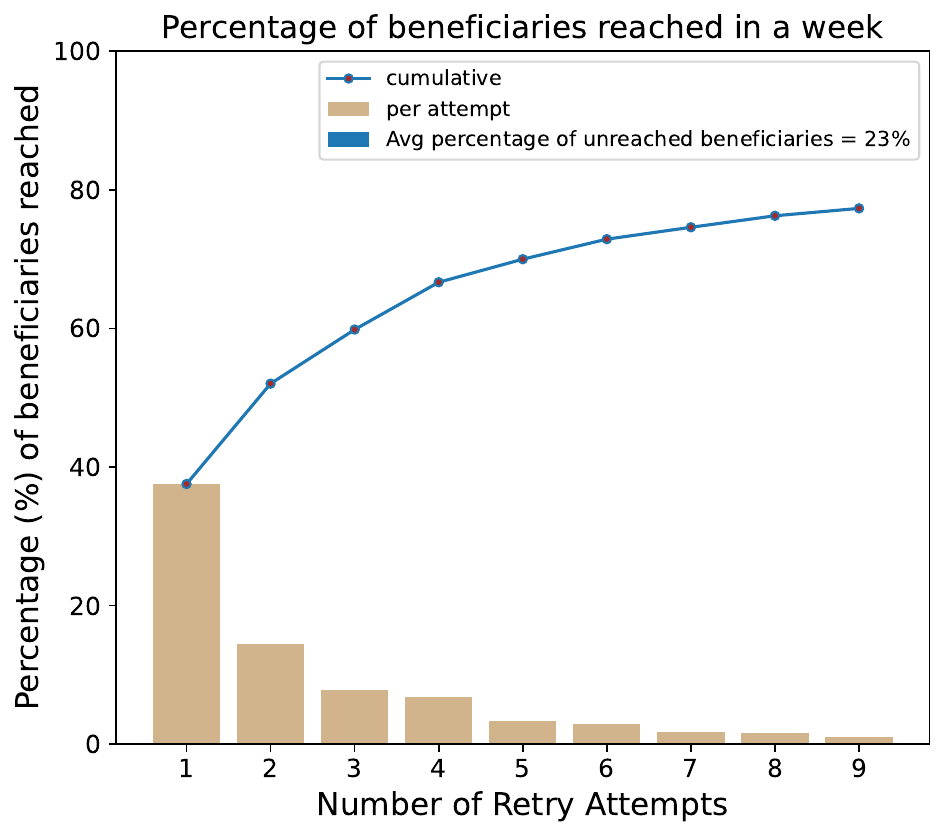}
  \caption{Efficacy of each successive attempt}
  \label{fig:att1}
\end{figure}

The IVR System makes a maximum of 9 attempts to a beneficiary in a week. %
Fig.\ref{fig:att1} shows the percentage of beneficiaries reached with each successive attempt. The magnitude of the bars indicates the effectiveness of an attempt in reaching beneficiaries. We see a steady decrease in the percentage of beneficiaries that are reached with each successive attempt. The blue line indicates the cumulative percentage of beneficiaries reached. It can be seen that on average, 23\% beneficiaries are not reached despite 9 attempts. Given the scale of the program, this is a significant number.

\subsection{Analysing Listenership Patterns of Beneficiary buckets}
Based on call pickup and call engagement behaviour, the beneficiaries were divided into four buckets: High Pickup Rate - High Engagement Rate, High Pickup Rate - Low Engagement Rate, Low Pickup Rate - High Engagement Rate, Low Pickup Rate - Low Engagement Rate. To avoid noisy patterns caused due to ad hoc listenership, two extreme set of beneficiaries were picked from the original data - those that have a high listening trajectory of greater than 50 weeks and those who have a lower listening trajectory of less than 20 weeks. These sets were constructed for the purpose of preliminary analysis, and we plan to expand them further in the future to include more beneficiaries. The previously mentioned four buckets were then constructed from the union of these two sets of beneficiaries. The distribution of beneficiaries per bucket is as per Table \ref{tab:buckets}.
\begin{table}
\centering
\begin{tabular}{| l | l | l |}
 \hline
\diagbox[trim=l,height=1.5\line]{Pickup}{Listenership} & High & Low    \\
 \hline
High                                          & 1212 & 2651   \\
 \hline
Low                                           & 1421 & 6087    \\
 \hline                                    
\end{tabular}
\caption{\label{tab:buckets} Distribution of beneficiary in each bucket}
\end{table}

Three major conclusions can be drawn from the analysis:
\begin{itemize}

    \item High pickup rate need not translate to high engagement.
    \item Contacting beneficiaries in their preferred time slots could increase listenership.
    \item Technical failures reduce pickup rates and could potentially lead to dropouts.

\end{itemize}

\begin{figure}[h!]
  \includegraphics[width=\linewidth]{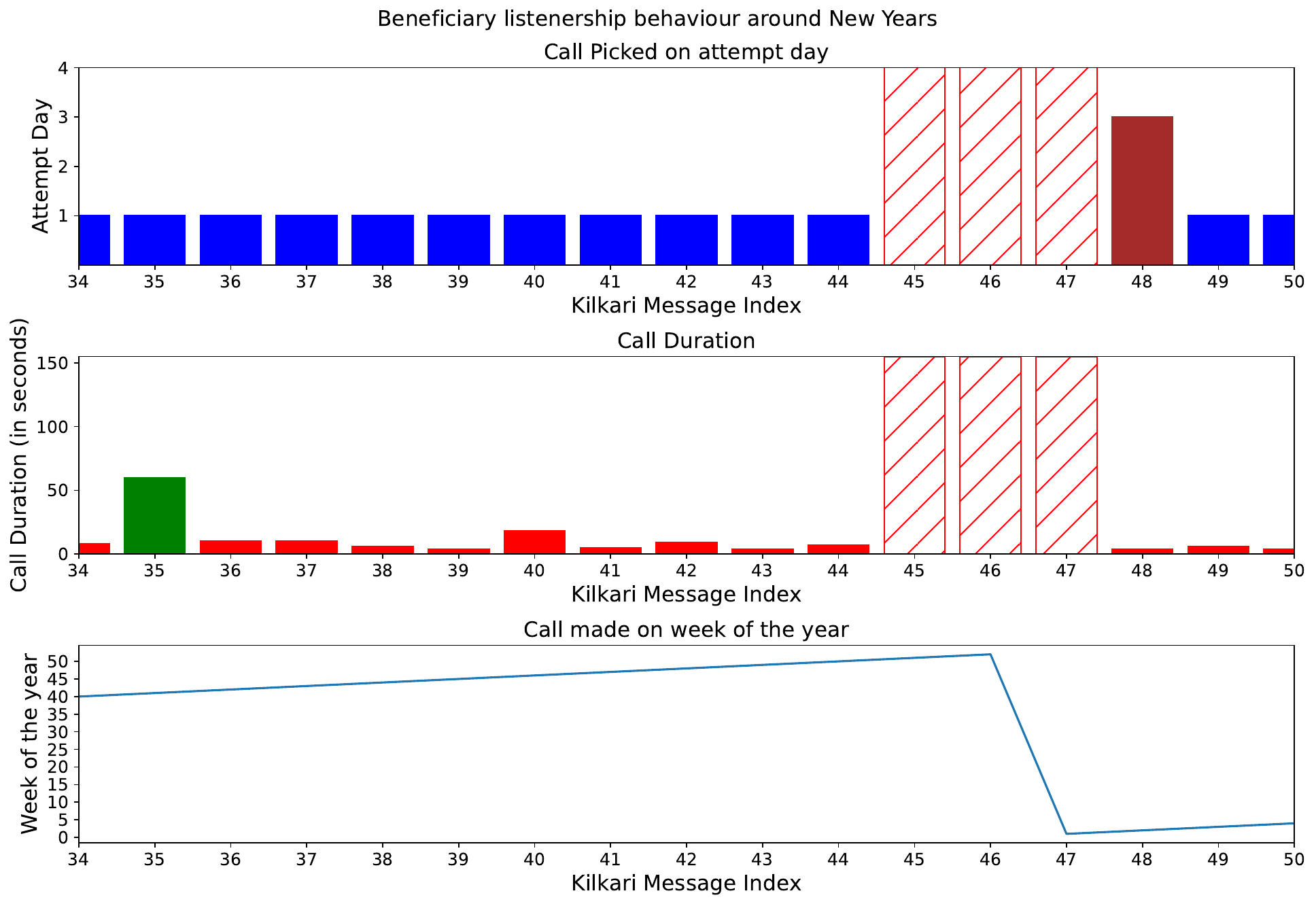}
  \caption{Listenership pattern of a beneficiary around new year - the red hashed bars indicate that the beneficiary did not pick up calls around new years }
  \label{fig:ny}
\end{figure}

Some external factors could also affect beneficiary pick up rates and engagement. For example, there was an observed decrease in listenership of specific beneficiaries during the New Year period, which could be a possible phenomenon during most major festivities. Fig.\ref{fig:ny} illustrates this for one beneficiary. The first graph indicates which attempt day the call was picked on. The red hashed bars indicate weeks when the call was not picked up. The second plot shows whether the beneficiary engaged with the call (green bar) or not (red bar). Finally, the last graph indicates the week of the year at each Kilkari message index. We see the period of new year's starts around message index 45, around which the beneficiary doesn't pick up calls. 

For figures \ref{fig:hh}, \ref{fig:hl}, \ref{fig:lh} and \ref{fig:ll}, the six graphs display various characteristics of these aggregate groups. The first left graph shows the average pickup rate. The first right graph shows on average, what attempt day did the beneficiaries pick up the call in ('blue' if attempt day 1, 'green' if attempt day 2, 'brown' if attempt day 3, 'red' if attempt day 4). The second left graph indicates the average technical success ratio (a green dot indicates a technical success ratio greater than 0.5, a red dot will indicate otherwise). The second right graph indicates which time slot did most of the beneficiaries pick up in out of the 7 time slots . Time Slot 0 is the earliest time slot (8AM-10PM) and Time Slot 6 is the latest (8PM-10PM) The third left graph indicates the call duration on average for every message index ('red' i.e non-engaging if call duration $<=$ 30.0 seconds else 'green' i.e engaging). The third right graph indicates which day of the week did most of the beneficiaries pick up in. Day 0 indicates a Monday and Day 6 a Sunday. The color-coded scatter plots of time slot and day of the week are for easy visualisation of which points are in abundance.

\subsubsection{Bucket 1: High Pickup Rate - High Engagement Rate beneficiaries}

\begin{figure}[h!]
  \includegraphics[width=\linewidth]{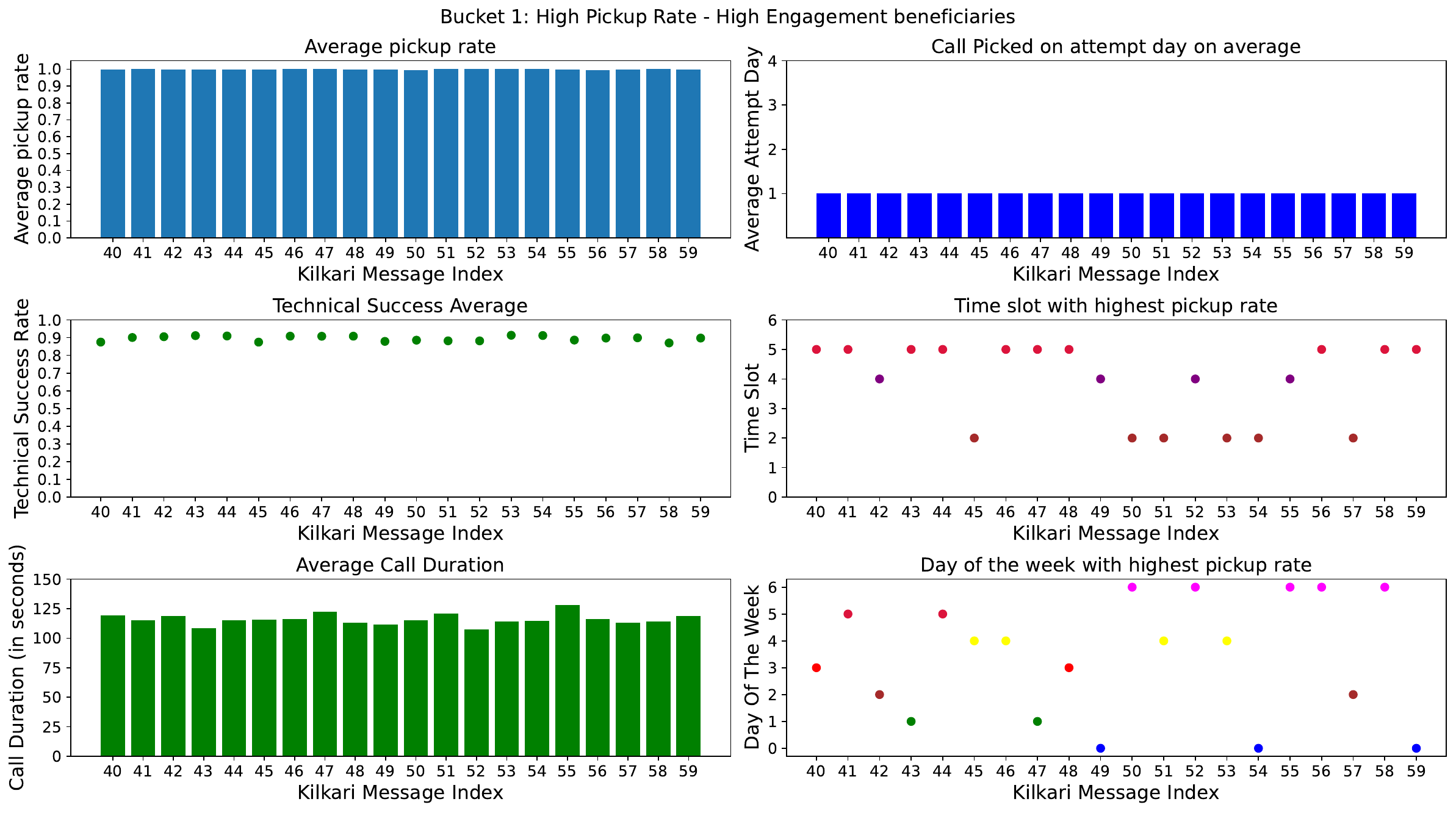}
  \caption{Listenership patterns of beneficiaries with high pick up rates and high listenership}
  \label{fig:hh}
\end{figure}

On average, this group of beneficiaries (Fig. \ref{fig:hh}) picked up the call on the first attempt day. The average technical success rate for these beneficiaries was extremely high, around 90\%. Thus, high pickup rates seem to be indicative of a good technical success ratio. 
Most of these beneficiaries seem to pick up across diverse call time slots, with a slight preference for time slot 5 (6 PM to 8 PM).

\subsubsection{Bucket 2: High Pickup Rate - Low Engagement Rate beneficiaries}

\begin{figure}[h!]
  \includegraphics[width=\linewidth]{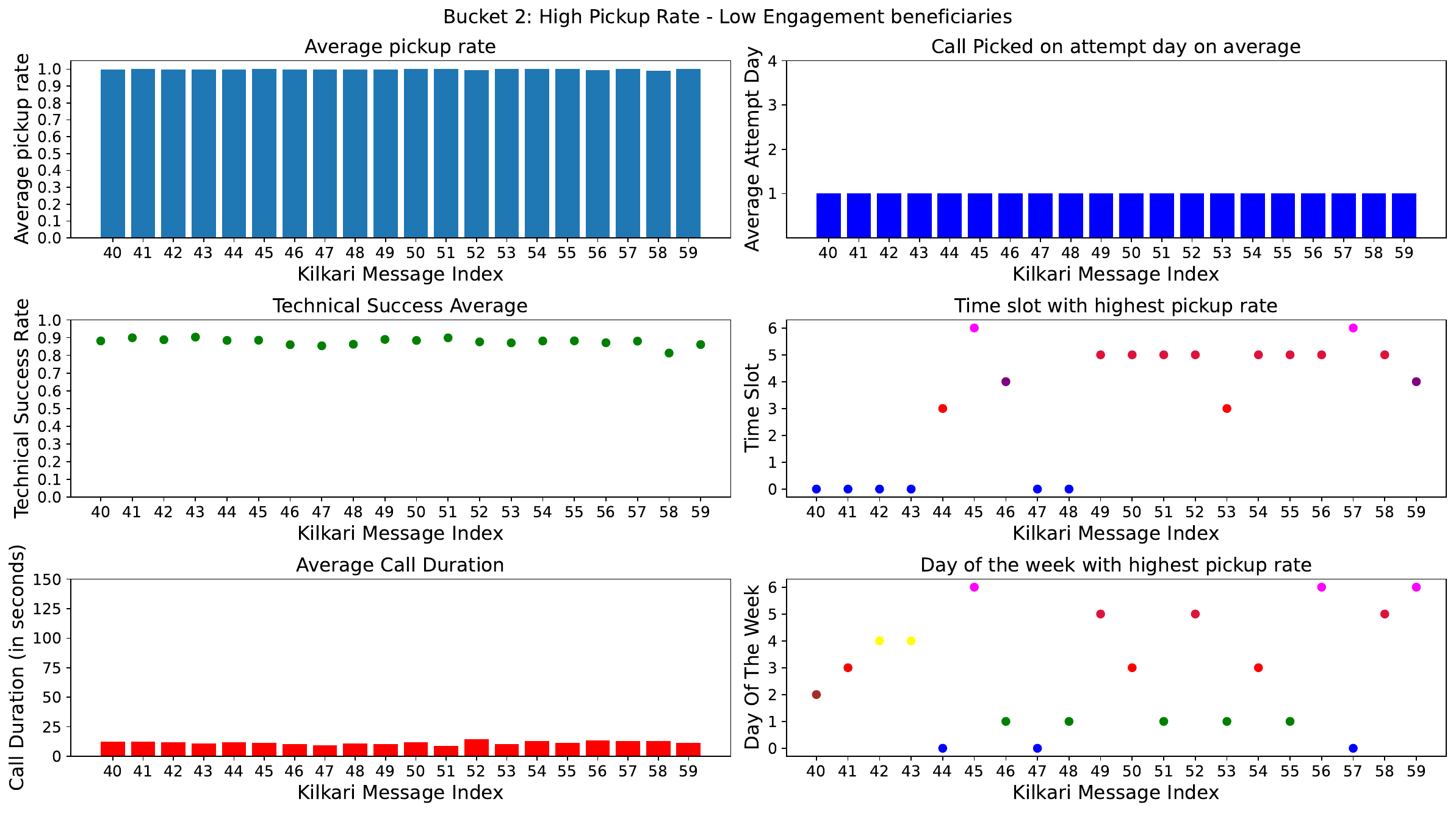}
  \caption{Listenership patterns of beneficiaries with high pick up rates and low listenership}
  \label{fig:hl}
\end{figure}

The second bucket (Fig.\ref{fig:hl}) comprises of beneficiaries that highlight one of the key focus points of this paper - \textbf{ engagement in the program is a better metric than pickup rates to measure program success}. Despite a high average technical success rate, the engagement in the program remains low. 
While a predictive model for pickup rates will prefer this set of beneficiaries over beneficiaries with lower pick up rates, a predictive model for call engagement will ensure that beneficiaries more receptive to the program are given their due attention.

\subsubsection{Bucket 3: Low Pickup Rate - High Engagement Rate beneficiaries}

\begin{figure}[h!]
  \includegraphics[width=\linewidth]{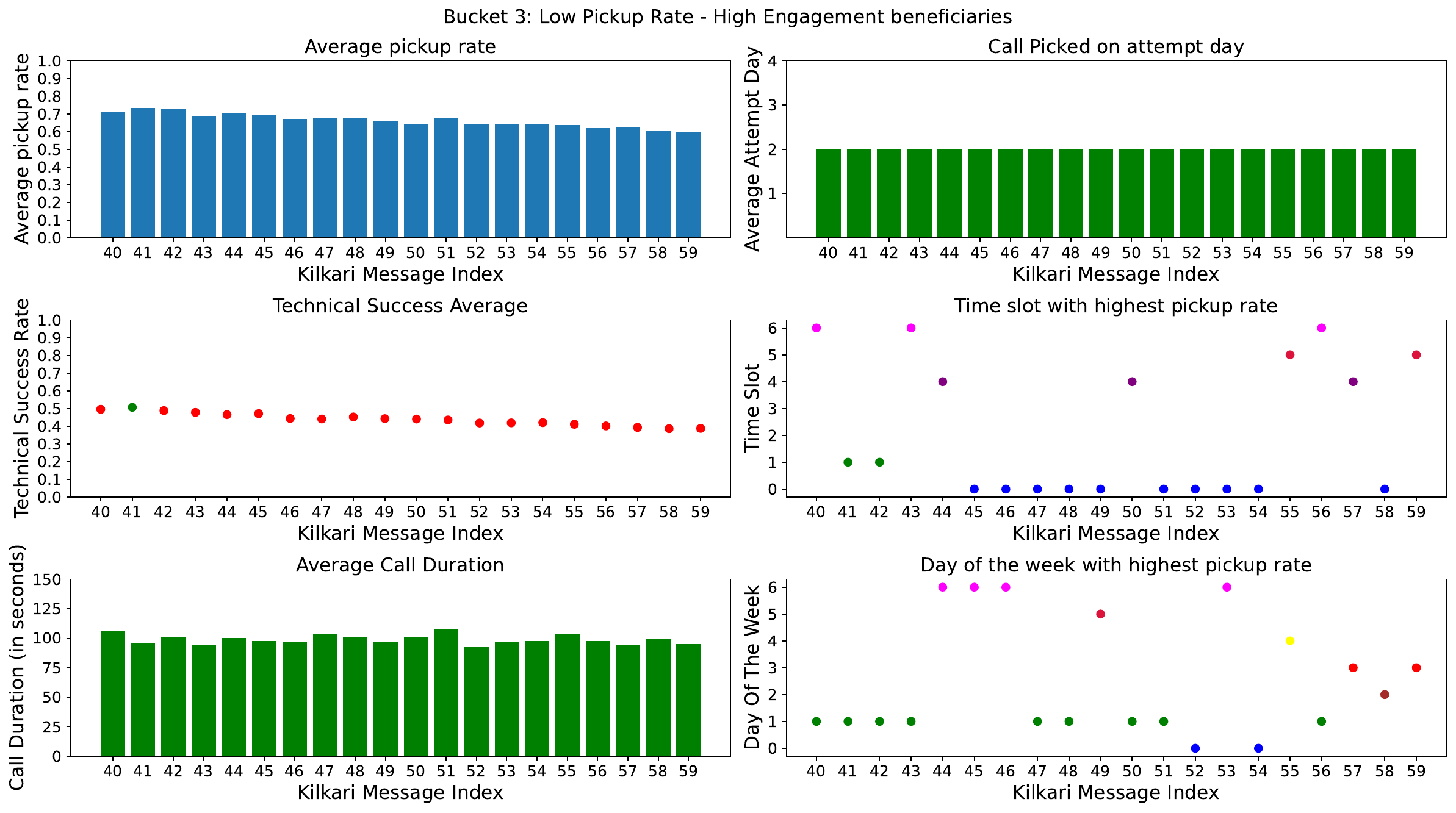}
  \caption{Listenership patterns of beneficiaries with low pick up rates and high listenership}
  \label{fig:lh}
\end{figure}

The third bucket (Fig.\ref{fig:lh}) of beneficiaries tends to highly engage with the program, despite lower pickup rates.  A possible reason for lower pick up rates could be the low average technical success rates of calls. These beneficiaries also display a strong preference for the morning and evening slots, with most calls being picked up during the first two and last two slots of the day. Thus, providing a higher number of attempts to address technical failures and rearranging calls for beneficiaries in this bucket for early or late in the day can potentially improve their performance in the program .

\subsubsection{Bucket 4: Low Pickup Rate - Low Engagement Rate beneficiaries}

\begin{figure}[h!]
  \includegraphics[width=\linewidth]{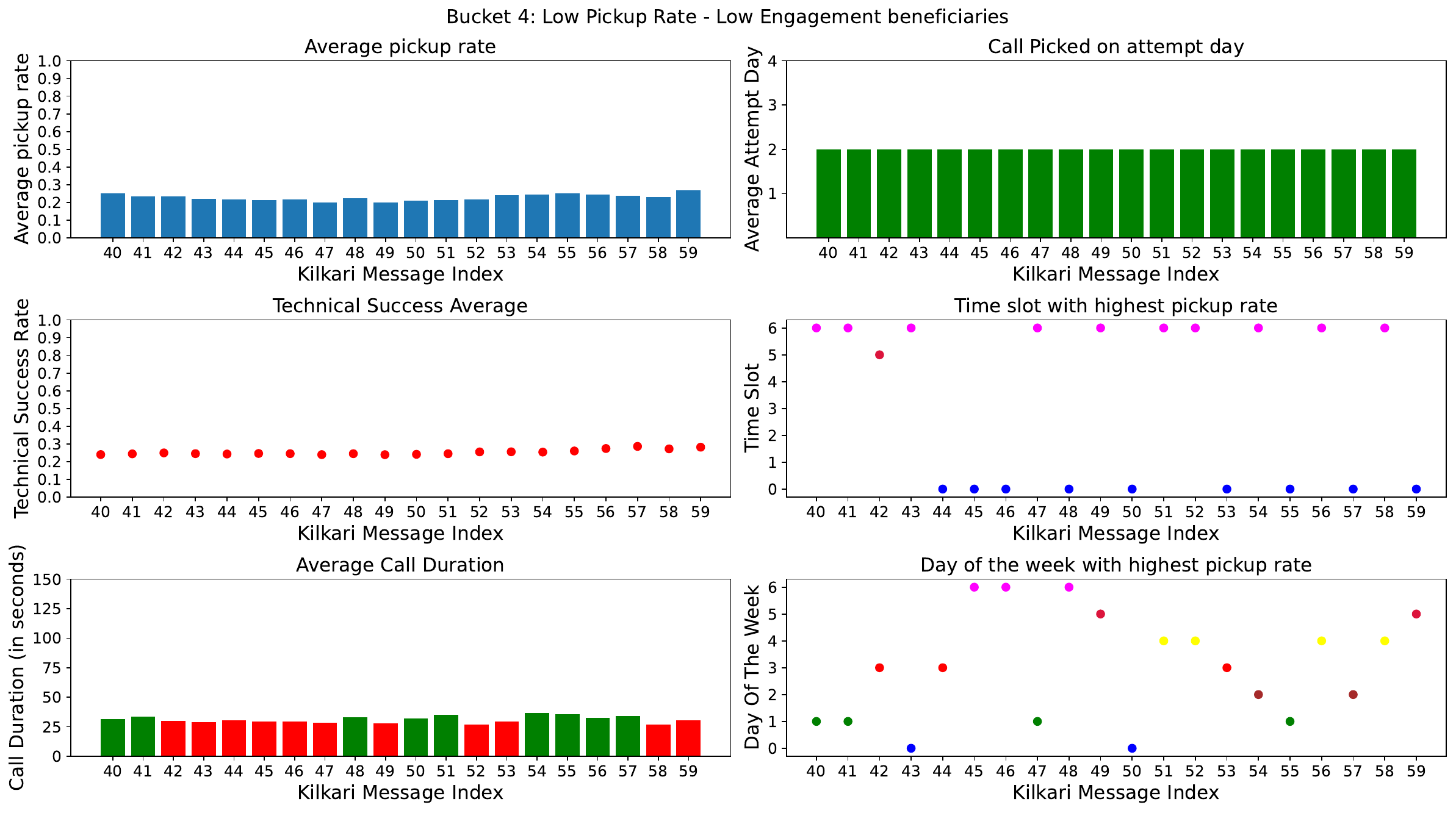}
  \caption{Listenership patterns of beneficiaries with low pick up rates and low listenership}
  \label{fig:ll}
\end{figure}

The average technical success rate for these beneficiaries (Fig. \ref{fig:ll}) is the lowest out of all the buckets, and is a meager 20\%. This indicates that most calls made to these beneficiaries do not go through. This could be contributory to their low pickup rates. Finally, the beneficiaries also show a preference for the first and last time slots of the day. 
Addressing the correct time slots to place calls, driving engagement for calls that do connect despite lower pick up rates, and increasing the number of attempts to counter the effect of technical failures can potentially boost the performance of these beneficiaries. 

\subsection{Technical Failures and pick up rates}

Certain beneficiaries are more susceptible to technical failures than others, which %
can affect their pickup rates %
and might eventually lead to them getting dropped out of the program. Dropouts are crucial in Kilkari as they save resources for the program by avoiding unnecessary attempts to phone numbers that are no longer in service, or to people who were incorrectly registered in the program. However, an unintended effect may be the possibility of dropouts in poor network coverage areas. While such issues can be addressed in due course with improvements in network coverage, it is critical to make multiple attempts to ensure maximum beneficiaries can receive technically successful calls.

\begin{figure}[h!]
  \includegraphics[width=\linewidth]{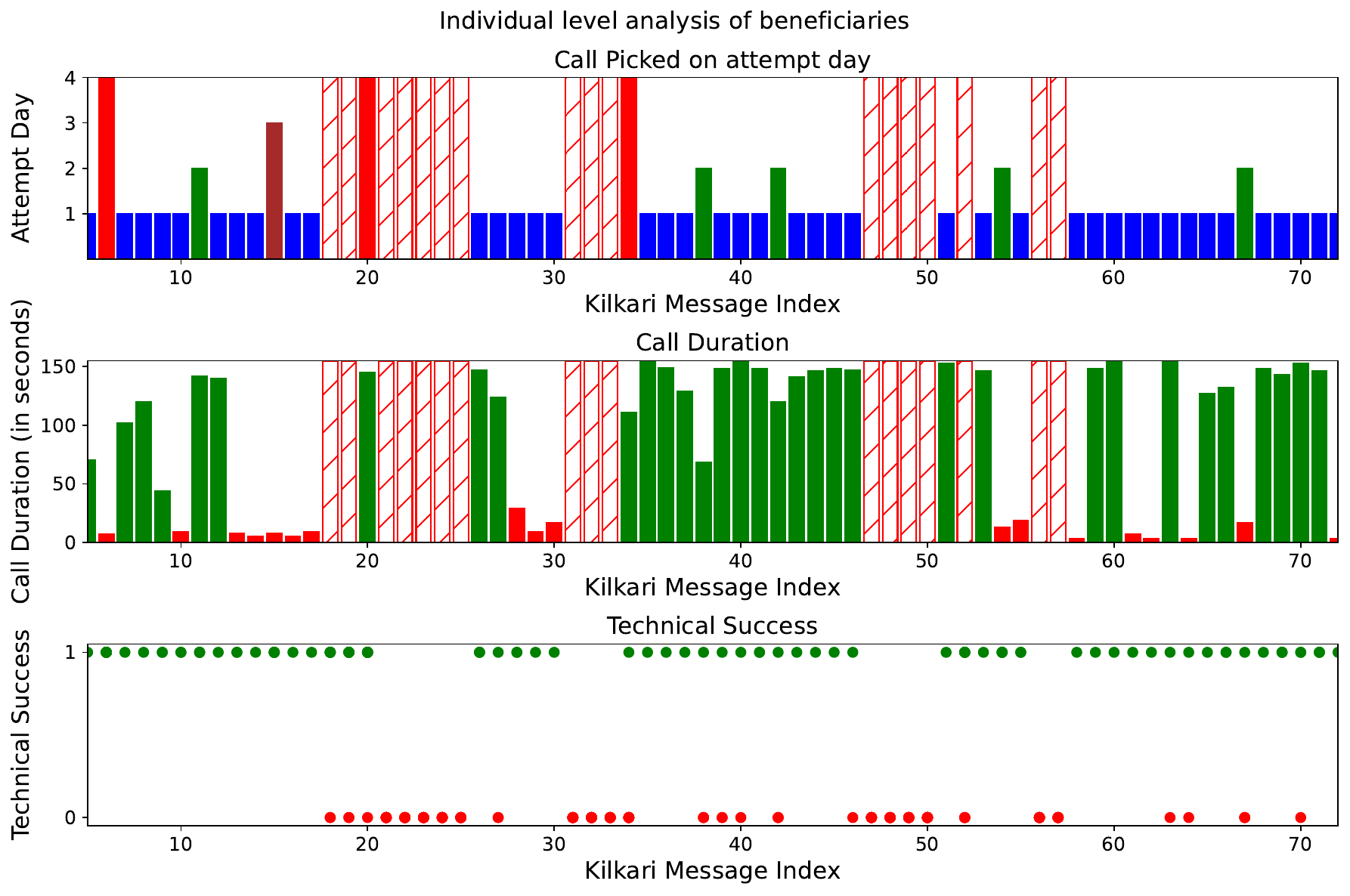}
  \caption{ A beneficiary who picks up all calls except when they are complete technical failures - the beneficiary loses upto 5 weeks of messages due to technical failures }
  \label{fig:tf2}
\end{figure}

Fig.\ref{fig:tf2} shows the listening trajectory of a beneficiary who picks up all the calls. The first graph shows the attempt day on which the call was picked ('blue' if attempt day = 1, 'green' if attempt day = 2, 'maroon' if attempt day = 3, 'red' if attempt day = 4). The second graph shows the engagement of the beneficiary every week (green bar implies engaging and red bar implies non-engaging). The red hashed bars in both the graphs indicate calls that haven't been picked up. Finally, the final graph plots Boolean values of whether the beneficiary received technically successful (green) and failed (red) calls. Weeks with only red dots indicate that no technically successful calls were made that week. %

We see that the %
beneficiary loses up to 5 weeks of messages due to a series of technical failures. It is interesting to note that after 6 weeks of no listenership, beneficiaries are dropped from the program.%
Kilkari has the flexibility of increasing the number of attempt days that beneficiaries are contacted over, hence identifying the segment of beneficiaries prone to technical failures and reaching them over a greater number of attempts can help further boost pickup rates in the program.

\subsection{Analysis of time slot on listener pickup and engagement rates}

\begin{figure}[h!]
\includegraphics[scale=0.5]{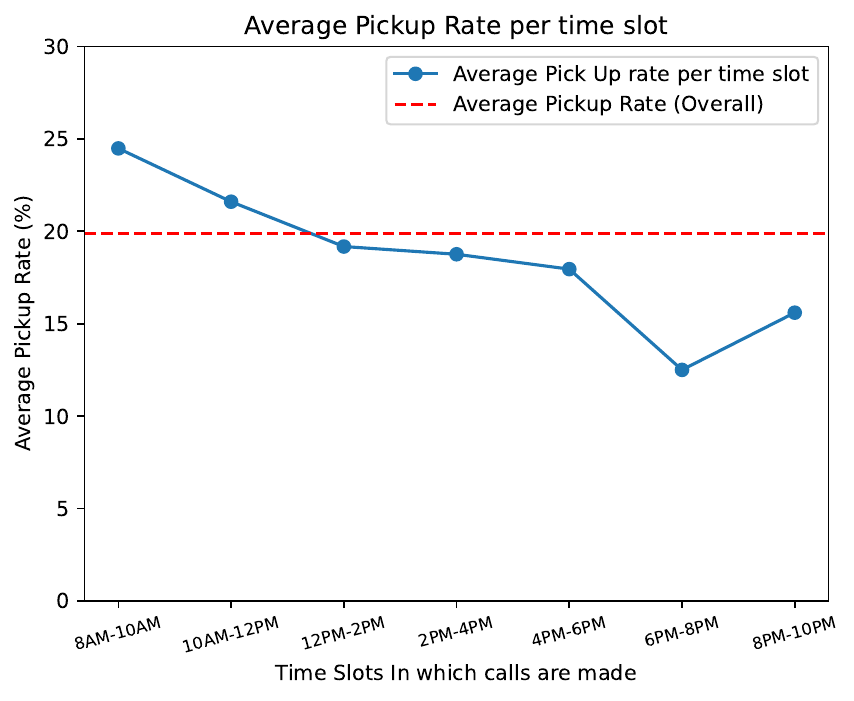}
  \caption{ Pickup rates per time slot for the full cohort }
  \label{fig:tslot1}
\end{figure}

\begin{figure}[h!]
  \includegraphics[scale=0.5]{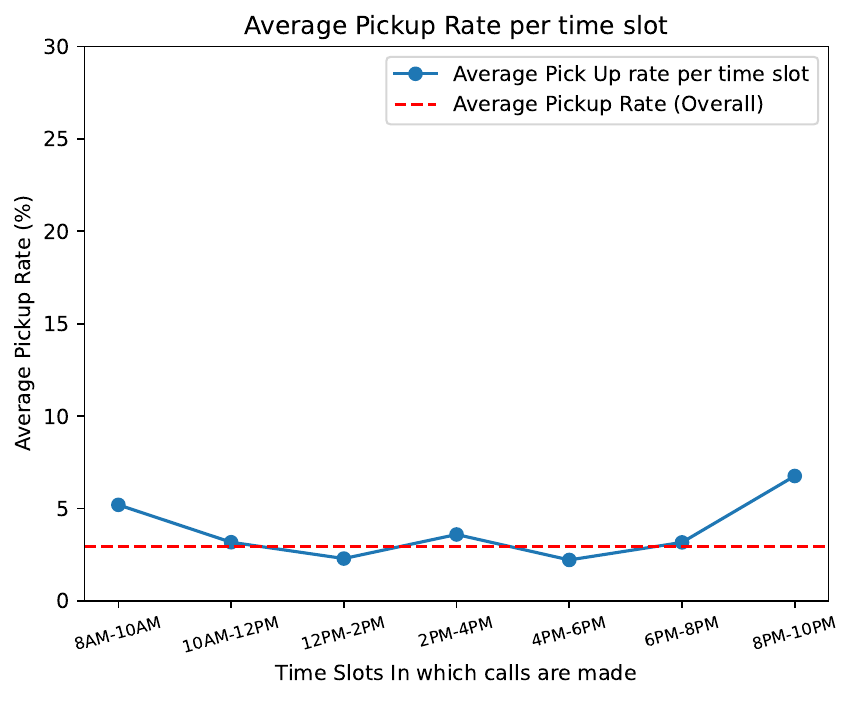}
  \caption{ Pickup rates per time slot for bucket 4 }
  \label{fig:tslot2}
\end{figure}

Currently, calls in Kilkari are made at random times during the day. For analysis, we have constructed seven time slots in the interval from 8 AM to 10 PM, divided into two-hour intervals. Calling beneficiaries in their preferred time slot can help in increasing pickup and engagement rates. As seen in Fig.\ref{fig:tslot1}, there is a clear preference in the total cohort for morning time slots, which boasts of the highest pickup rates. However, for %
beneficiaries with low pickup rates and low engagement, all slots perform equally poorly with the exception of the first and last time slot (Fig.\ref{fig:tslot2}). As discussed previously, one possible reason is the issue of phone ownership, where the beneficiary has access to the phone in only the earlier or later parts of the day. \\

Currently, there are plans to update the IVR infrastructure to allow for maximum outreach. While bandwidth will no longer be an issue in the newer version of the IVR system, concurrency of calls still acts as a bottleneck. Thus, it might not be possible to call all beneficiaries in the morning or evening slots and leave the remaining slots with no bandwidth utilization. We aim to build a predictive model to calculate the top-k preferred time slots of beneficiaries, and optimize calling times in a way that can bring up the overall pickup and engagement rates of the cohort while working under the constraint of the number of concurrent calls that can be made at a particular time. 

\begin{table*}[ht]
\centering
\begin{tabular}{@{}ccccc@{}}
\toprule
\multicolumn{1}{c}{Model} & \multicolumn{1}{c}{Features} & \multicolumn{1}{c}{Balanced Accuracy} & \multicolumn{1}{c}{Precision@5} & \multicolumn{1}{c}{AUC} \\ \midrule
Logistic Regression                    & duration, attempt, status, date      & 0.756635                              & 0.871566                        & 0.828824                \\
Logistic Regression                    & duration, attempt, status           & 0.756471                              & 0.869925                        & 0.828644                \\
Feedforward NN                     & duration, attempt, status           & 0.756136                              & 0.872381                        & 0.828275                \\
Feedforward NN                     & duration, attempt                  & 0.756120                              & 0.872837                        & 0.828449                \\
LSTM                      & duration, attempt                  & 0.756097                              & 0.876322                        & 0.828866                \\
Feedforward NN                     & duration, attempt, status, date      & 0.755964                              & 0.871730                        & 0.828090                \\
Logistic Regression                    & duration, attempt                  & 0.755885                              & 0.874805                        & 0.828331                \\
LSTM                      & duration, attempt, status, date      & 0.755800                              & 0.866317                        & 0.827406                \\
LSTM                      & duration, attempt, status           & 0.754932                              & 0.865422                        & 0.826992                \\
Random                    & duration, attempt, status, date      & 0.501689                              & 0.410071                        & 0.501652                \\
Random                    & duration, attempt                  & 0.501505                              & 0.409128                        & 0.501464                \\
Random                    & duration, attempt, status           & 0.500297                              & 0.408431                        & 0.498771                \\ \bottomrule
\end{tabular}
\caption{Overall Results in predictive modeling for low-engagement prediction}
\label{tab:low-eng}
\end{table*}

\begin{table*}[ht]
\centering
\begin{tabular}{@{}ccccc@{}}
\toprule
\multicolumn{1}{c}{Model} & \multicolumn{1}{c}{Features} & \multicolumn{1}{c}{Balanced Accuracy} & \multicolumn{1}{c}{Precision@5} & \multicolumn{1}{c}{AUC} \\ \midrule
Feedforward NN                     & duration, attempt, status, date      & 0.797638                              & 0.430001                        & 0.872797                \\
Logistic Regression                    & duration, attempt, status, date      & 0.796492                              & 0.424793                        & 0.870911                \\
Feedforward NN                     & duration, attempt, status           & 0.796021                              & 0.425285                        & 0.870253                \\
Logistic Regression                    & duration, attempt, status           & 0.795685                              & 0.422127                        & 0.870115                \\
Feedforward NN                     & duration, attempt                  & 0.795241                              & 0.420856                        & 0.869499                \\
Logistic Regression                    & duration, attempt                  & 0.794780                              & 0.416838                        & 0.868858                \\
LSTM                      & duration, attempt                  & 0.794487                              & 0.421922                        & 0.869370                \\
Random                    & duration, attempt                  & 0.503113                              & 0.055565                        & 0.503159                \\
Random                    & duration, attempt, status, date      & 0.502418                              & 0.053555                        & 0.500930                \\
Random                    & duration, attempt, status           & 0.501334                              & 0.055196                        & 0.499087                           \\ \bottomrule
\end{tabular}
\caption{Overall Results in predictive modeling for low-call pickup prediction}
\label{tab:low-pick}
\end{table*}

\subsection{Predicting low-listeners}

To segment beneficiaries into different clusters and to help NGOs plan for timely interventions that can increase program retention, we make use of beneficiaries' historical trajectories in a time-series model to predict low pickup rates and low engagement rates. Our model does not make use of any beneficiary demographic features, and is thus especially relevant for mHealth programs, where beneficary information is often sensitive and limited. 

We use the dataset described in Section 3.3 to predict whether beneficiaries would showcase low-listenership behaviour in the future. Specifically, we consider two definitions of low-listenership, low-pickup and low engagement. We thus learn separate binary classification models for these two target variables. As input, we consider $N_{train}=6$ weeks of listenership indicators. This includes time series of duration of calls listened to every week, number of call attempts every week and number of calls with every status codes every week. We consider the following status codes that is logged in the Kilkari system: call picked up, phone busy, phone switched off, phone out of network, any other reason of not reaching. This results in a total of 7 features every week. Crucially, we develop predictive models which only use listenership behaviour data and no beneficiary specific information.

We consider the following models in our experiments:

\begin{itemize}
    \item \textbf{Random}: This is the baseline model which predicts low listenership by sampling from uniform random distribution
    \item \textbf{Logistic Regression}: This uses a logistic regression model to predict the target variable. Time series features are flattened and given as input to the model
    \item \textbf{Feedforward NN}: This uses a dense feedforward neural network. We use 3 layers with 128 hidden units each and finally a sigmoid acitvation for output. Binary crossentropy loss is used to optimise the weights of the neural network. This model also uses flattened time series as input.
    \item \textbf{LSTM}: We use Keras implementation of Long Short Term Memory model for encoding sequential information from beneficiaires listenership. Specifically, we use LSTM cell with 128 hidden units followed by three layers of 128 hidden units each. Finally, we use sigmoid activation for output. and Binary crossentropy loss as optimization objective.
\end{itemize}

\noindent We evaluate all models on three metrics:
\begin{itemize}
    \item \textbf{Precision@K\%:} This is the precision in predicting low-listenership flag if we set threshold to $(100-K)^{\text{th}}$ percentile of values. This measures the fraction of low-listeners that we will find if top $K\%$ of beneficiaires are chosen according to model predictions. Since health resources are limited, we use ($K=5\%$) in our experiments.
    \item \textbf{Balanced Accuracy:} Since we have imbalanced target distribution, we extend the notion of accuracy to imbalanced classes. Balanced accuracy is the arithmetic mean of sensitivity and specificity. 
    \item \textbf{AUC}: We also report the area under the ROC curve (AUC) metric to compare all the models.
\end{itemize}

In table \ref{tab:low-eng} and table \ref{tab:low-pick}, we compare different models on multiple evaluation metrics for low-engagement and low-pickup targets respectively. We notice for both the targets, ML models can perform much better than random. This showcases that input features are predictive of low listenership. However, we do not see any added value from time series model like LSTM, as opposed to logistic regression or feedforward neural network. Lastly, we find that adding additional features, such as status of calls and attempts results in slight increase in predictive performance.

\section{Ethics and Data Usage}
Acknowledging the responsibility associated with real-world AI systems for undeserved communities, we have closely coordinated with domain experts from the NGO throughout our analysis. This study falls into the category of secondary analysis of the aforementioned dataset. We use the previously collected engagement trajectories of different beneficiaries participating in the service call program to train the predictive model and evaluate the performance.
All the data collected through the program is owned by the NGO and only the NGO is allowed to share data while the research group only accesses an anonymized version of the data.

\section{Conclusion}
In this preliminary work, we perform secondary data analysis for Kilkari, the largest maternal mobile health program in the world. We characterize beneficiaries' engagement and pickup behaviour with the program through their temporal patterns in voice message listenership.
We showcase that apart from beneficiaries' call pickup rates, which has been the primary focus of several previous works, call engagement rates, preferred time slots and good call technical success ratios are also critical for successful outreach of the program. Lastly, we demonstrate that historical data can be used to predict low-listenership of beneficiaries to provide tailored approaches for beneficiary clusters and help NGOs perform timely intervention to increase beneficiary retention. Our proposed modelling approach relies only on past listenership trajectories, thus removing dependence on any sensitive or limited beneficiary information. These results also open up new line of ML research in areas such as feature engineering, intervention optimization and measuring behaviour change with better access to information such as voice messages. Our analysis and techniques can be extended to other large-scale mHealth programs.

\bibliographystyle{ACM-Reference-Format}
\bibliography{sample-base}


\begin{thebibliography}{18}

%%% the Web address.

\ifx \showCODEN    \undefined \def \showCODEN     #1{\unskip}     \fi
\ifx \showDOI      \undefined \def \showDOI       #1{#1}\fi
\ifx \showISBNx    \undefined \def \showISBNx     #1{\unskip}     \fi
\ifx \showISBNxiii \undefined \def \showISBNxiii  #1{\unskip}     \fi
\ifx \showISSN     \undefined \def \showISSN      #1{\unskip}     \fi
\ifx \showLCCN     \undefined \def \showLCCN      #1{\unskip}     \fi
\ifx \shownote     \undefined \def \shownote      #1{#1}          \fi
\ifx \showarticletitle \undefined \def \showarticletitle #1{#1}   \fi
\ifx \showURL      \undefined \def \showURL       {\relax}        \fi
% invisible to TeX
\providecommand\bibfield[2]{#2}
\providecommand\bibinfo[2]{#2}
\providecommand\natexlab[1]{#1}
\providecommand\showeprint[2][]{arXiv:#2}

\bibitem[Aranda-Jan et~al\mbox{.}(2014)]%
        {aranda2014systematic}
\bibfield{author}{\bibinfo{person}{Clara~B Aranda-Jan}, \bibinfo{person}{Neo
  Mohutsiwa-Dibe}, {and} \bibinfo{person}{Svetla Loukanova}.}
  \bibinfo{year}{2014}\natexlab{}.
\newblock \showarticletitle{Systematic review on what works, what does not work
  and why of implementation of mobile health (mHealth) projects in Africa}.
\newblock \bibinfo{journal}{\emph{BMC public health}} \bibinfo{volume}{14},
  \bibinfo{number}{1} (\bibinfo{year}{2014}), \bibinfo{pages}{1--15}.
\newblock


\bibitem[{ARMMAN}({[n.\,d.]})]%
        {KilkariArticle}
\bibfield{author}{\bibinfo{person}{{ARMMAN}}.}
  \bibinfo{year}{[n.\,d.]}\natexlab{}.
\newblock \bibinfo{title}{Kilkari}.
\newblock
\newblock
\urldef\tempurl%
\url{https://armman.org/kilkari/}
\showURL{%
\tempurl}
\newblock
\shownote{Accessed on April 5, 2023}.


\bibitem[{ARMMAN}(2019)]%
        {armman-mhealth}
\bibfield{author}{\bibinfo{person}{{ARMMAN}}.} \bibinfo{year}{2019}\natexlab{}.
\newblock \bibinfo{title}{Assessing the Impact of Mobile-based Intervention on
  Health Literacy among Pregnant Women in Urban India}.
\newblock
  \bibinfo{howpublished}{\url{https://armman.org/wp-content/uploads/2019/09/Sion-Study-Abstract.pdf}}.
\newblock
\newblock
\shownote{Accessed: 2022-08-12}.


\bibitem[Chamberlain et~al\mbox{.}(2021)]%
        {chamberlain2021ten}
\bibfield{author}{\bibinfo{person}{Sara Chamberlain}, \bibinfo{person}{Priyanka
  Dutt}, \bibinfo{person}{Anna Godfrey}, \bibinfo{person}{Radharani Mitra},
  \bibinfo{person}{Amnesty~Elizabeth LeFevre}, \bibinfo{person}{Kerry Scott},
  \bibinfo{person}{Jai Mendiratta}, \bibinfo{person}{Vinod Chauhan}, {and}
  \bibinfo{person}{Salil Arora}.} \bibinfo{year}{2021}\natexlab{}.
\newblock \showarticletitle{Ten lessons learnt: scaling and transitioning one
  of the largest mobile health communication programmes in the world to a
  national government}.
\newblock \bibinfo{journal}{\emph{BMJ Global Health}} \bibinfo{volume}{6},
  \bibinfo{number}{Suppl 5} (\bibinfo{year}{2021}), \bibinfo{pages}{e005341}.
\newblock


\bibitem[Guo et~al\mbox{.}(2020)]%
        {guo2020mobile}
\bibfield{author}{\bibinfo{person}{Yutao Guo}, \bibinfo{person}{Deirdre~A
  Lane}, \bibinfo{person}{Limin Wang}, \bibinfo{person}{Hui Zhang},
  \bibinfo{person}{Hao Wang}, \bibinfo{person}{Wei Zhang},
  \bibinfo{person}{Jing Wen}, \bibinfo{person}{Yunli Xing},
  \bibinfo{person}{Fang Wu}, \bibinfo{person}{Yunlong Xia}, {et~al\mbox{.}}}
  \bibinfo{year}{2020}\natexlab{}.
\newblock \showarticletitle{Mobile health technology to improve care for
  patients with atrial fibrillation}.
\newblock \bibinfo{journal}{\emph{Journal of the American College of
  Cardiology}} \bibinfo{volume}{75}, \bibinfo{number}{13}
  (\bibinfo{year}{2020}), \bibinfo{pages}{1523--1534}.
\newblock


\bibitem[Killian et~al\mbox{.}(2019)]%
        {Killian_2019}
\bibfield{author}{\bibinfo{person}{Jackson~A. Killian}, \bibinfo{person}{Bryan
  Wilder}, \bibinfo{person}{Amit Sharma}, \bibinfo{person}{Vinod Choudhary},
  \bibinfo{person}{Bistra Dilkina}, {and} \bibinfo{person}{Milind Tambe}.}
  \bibinfo{year}{2019}\natexlab{}.
\newblock \showarticletitle{Learning to Prescribe Interventions for
  Tuberculosis Patients Using Digital Adherence Data}.
\newblock \bibinfo{journal}{\emph{Proceedings of the 25th ACM SIGKDD
  International Conference on Knowledge Discovery \& Data Mining}}
  (\bibinfo{date}{Jul} \bibinfo{year}{2019}).
\newblock


\bibitem[Madanian et~al\mbox{.}(2019)]%
        {madanian2019mhealth}
\bibfield{author}{\bibinfo{person}{Samaneh Madanian}, \bibinfo{person}{Dave~T
  Parry}, \bibinfo{person}{David Airehrour}, {and} \bibinfo{person}{Marianne
  Cherrington}.} \bibinfo{year}{2019}\natexlab{}.
\newblock \showarticletitle{mHealth and big-data integration: promises for
  healthcare system in India}.
\newblock \bibinfo{journal}{\emph{BMJ health \& care informatics}}
  \bibinfo{volume}{26}, \bibinfo{number}{1} (\bibinfo{year}{2019}).
\newblock


\bibitem[Mate et~al\mbox{.}(2022)]%
        {mate2022field}
\bibfield{author}{\bibinfo{person}{Aditya Mate}, \bibinfo{person}{Lovish
  Madaan}, \bibinfo{person}{Aparna Taneja}, \bibinfo{person}{Neha Madhiwalla},
  \bibinfo{person}{Shresth Verma}, \bibinfo{person}{Gargi Singh},
  \bibinfo{person}{Aparna Hegde}, \bibinfo{person}{Pradeep Varakantham}, {and}
  \bibinfo{person}{Milind Tambe}.} \bibinfo{year}{2022}\natexlab{}.
\newblock \showarticletitle{Field study in deploying restless multi-armed
  bandits: Assisting non-profits in improving maternal and child health}. In
  \bibinfo{booktitle}{\emph{Proceedings of the AAAI Conference on Artificial
  Intelligence}}, Vol.~\bibinfo{volume}{36}. \bibinfo{pages}{12017--12025}.
\newblock


\bibitem[McCool et~al\mbox{.}(2022)]%
        {mccool2022mobile}
\bibfield{author}{\bibinfo{person}{Judith McCool}, \bibinfo{person}{Rosie
  Dobson}, \bibinfo{person}{Robyn Whittaker}, {and} \bibinfo{person}{Chris
  Paton}.} \bibinfo{year}{2022}\natexlab{}.
\newblock \showarticletitle{Mobile health (mHealth) in low-and middle-income
  countries}.
\newblock \bibinfo{journal}{\emph{Annual Review of Public Health}}
  \bibinfo{volume}{43} (\bibinfo{year}{2022}), \bibinfo{pages}{525--539}.
\newblock


\bibitem[Mohan et~al\mbox{.}(2022)]%
        {mohan2022optimising}
\bibfield{author}{\bibinfo{person}{Diwakar Mohan}, \bibinfo{person}{Jean
  Juste~Harrisson Bashingwa}, \bibinfo{person}{Kerry Scott},
  \bibinfo{person}{Salil Arora}, \bibinfo{person}{Sai Rahul},
  \bibinfo{person}{Nicola Mulder}, \bibinfo{person}{Sara Chamberlain}, {and}
  \bibinfo{person}{Amnesty~Elizabeth LeFevre}.}
  \bibinfo{year}{2022}\natexlab{}.
\newblock \showarticletitle{Optimising the reach of mobile health messaging
  programmes: an analysis of system generated data for the Kilkari programme
  across 13 states in India}.
\newblock \bibinfo{journal}{\emph{BMJ Global Health}} \bibinfo{volume}{6},
  \bibinfo{number}{Suppl 5} (\bibinfo{year}{2022}), \bibinfo{pages}{e009395}.
\newblock


\bibitem[Mohan et~al\mbox{.}(2021)]%
        {mohan2021can}
\bibfield{author}{\bibinfo{person}{Diwakar Mohan}, \bibinfo{person}{Kerry
  Scott}, \bibinfo{person}{Neha Shah}, \bibinfo{person}{Jean Juste~Harrisson
  Bashingwa}, \bibinfo{person}{Arpita Chakraborty}, \bibinfo{person}{Osama
  Ummer}, \bibinfo{person}{Anna Godfrey}, \bibinfo{person}{Priyanka Dutt},
  \bibinfo{person}{Sara Chamberlain}, {and} \bibinfo{person}{Amnesty~Elizabeth
  LeFevre}.} \bibinfo{year}{2021}\natexlab{}.
\newblock \showarticletitle{Can health information through mobile phones close
  the divide in health behaviours among the marginalised? An equity analysis of
  Kilkari in Madhya Pradesh, India}.
\newblock \bibinfo{journal}{\emph{BMJ Global Health}} \bibinfo{volume}{6},
  \bibinfo{number}{Suppl 5} (\bibinfo{year}{2021}), \bibinfo{pages}{e005512}.
\newblock


\bibitem[Nishtala et~al\mbox{.}(2021)]%
        {nishtala2021selective}
\bibfield{author}{\bibinfo{person}{Siddharth Nishtala}, \bibinfo{person}{Lovish
  Madaan}, \bibinfo{person}{Aditya Mate}, \bibinfo{person}{Harshavardhan
  Kamarthi}, \bibinfo{person}{Anirudh Grama}, \bibinfo{person}{Divy Thakkar},
  \bibinfo{person}{Dhyanesh Narayanan}, \bibinfo{person}{Suresh Chaudhary},
  \bibinfo{person}{Neha Madhiwalla}, \bibinfo{person}{Ramesh Padmanabhan},
  {et~al\mbox{.}}} \bibinfo{year}{2021}\natexlab{}.
\newblock \showarticletitle{Selective Intervention Planning using Restless
  Multi-Armed Bandits to Improve Maternal and Child Health Outcomes}.
\newblock \bibinfo{journal}{\emph{arXiv preprint arXiv:2103.09052}}
  (\bibinfo{year}{2021}).
\newblock


\bibitem[Pilote et~al\mbox{.}(1996)]%
        {10.1001/archinte.1996.00440020063008}
\bibfield{author}{\bibinfo{person}{Louise Pilote},
  \bibinfo{person}{Jacqueline~P. Tulsky}, \bibinfo{person}{Andrew~R. Zolopa},
  \bibinfo{person}{Judith~A. Hahn}, \bibinfo{person}{Gisela~F. Schecter}, {and}
  \bibinfo{person}{Andrew~R. Moss}.} \bibinfo{year}{1996}\natexlab{}.
\newblock \showarticletitle{{Tuberculosis Prophylaxis in the Homeless: A Trial
  to Improve Adherence to Referral}}.
\newblock \bibinfo{journal}{\emph{Archives of Internal Medicine}}
  \bibinfo{volume}{156}, \bibinfo{number}{2} (\bibinfo{date}{01}
  \bibinfo{year}{1996}), \bibinfo{pages}{161--165}.
\newblock


\bibitem[Shah et~al\mbox{.}({[n.\,d.]})]%
        {shahpreliminary}
\bibfield{author}{\bibinfo{person}{Sanket Shah}, \bibinfo{person}{Shresth
  Verma}, \bibinfo{person}{Amrita Mahale}, \bibinfo{person}{Kumar~Madhu Sudan},
  \bibinfo{person}{Aparna Hegde}, \bibinfo{person}{Aparna Taneja}, {and}
  \bibinfo{person}{Milind Tambe}.} \bibinfo{year}{[n.\,d.]}\natexlab{}.
\newblock \showarticletitle{Preliminary Results in Low-Listenership Prediction
  in One of the Largest Mobile Health Programs in the World}.
\newblock  (\bibinfo{year}{[n.\,d.]}).
\newblock


\bibitem[Son et~al\mbox{.}(2010)]%
        {son2010application}
\bibfield{author}{\bibinfo{person}{Youn-Jung Son}, \bibinfo{person}{Hong-Gee
  Kim}, \bibinfo{person}{Eung-Hee Kim}, \bibinfo{person}{Sangsup Choi}, {and}
  \bibinfo{person}{Soo-Kyoung Lee}.} \bibinfo{year}{2010}\natexlab{}.
\newblock \showarticletitle{Application of support vector machine for
  prediction of medication adherence in heart failure patients}.
\newblock \bibinfo{journal}{\emph{Healthcare informatics research}}
  \bibinfo{volume}{16}, \bibinfo{number}{4} (\bibinfo{year}{2010}),
  \bibinfo{pages}{253--259}.
\newblock


\bibitem[Tuldrà et~al\mbox{.}(1999)]%
        {HIV}
\bibfield{author}{\bibinfo{person}{Albert Tuldrà}, \bibinfo{person}{Ma~José
  Ferrer}, \bibinfo{person}{Carmina~R. Fumaz}, \bibinfo{person}{Ramon Bayés},
  \bibinfo{person}{Roger Paredes}, \bibinfo{person}{David~M. Burger}, {and}
  \bibinfo{person}{Bonaventura Clotet}.} \bibinfo{year}{1999}\natexlab{}.
\newblock \showarticletitle{{Monitoring Adherence to HIV Therapy}}.
\newblock \bibinfo{journal}{\emph{Archives of Internal Medicine}}
  \bibinfo{volume}{159}, \bibinfo{number}{12} (\bibinfo{date}{06}
  \bibinfo{year}{1999}), \bibinfo{pages}{1376--1377}.
\newblock


\bibitem[Verma et~al\mbox{.}(2023)]%
        {51839}
\bibfield{author}{\bibinfo{person}{Shresth Verma}, \bibinfo{person}{Gargi
  Singh}, \bibinfo{person}{Aditya~S. Mate}, \bibinfo{person}{Paritosh Verma},
  \bibinfo{person}{Sruthi Gorantala}, \bibinfo{person}{Neha Madhiwalla},
  \bibinfo{person}{Aparna Hegde}, \bibinfo{person}{Divy~Hasmukhbhai Thakkar},
  \bibinfo{person}{Manish Jain}, \bibinfo{person}{Milind~Shashikant Tambe},
  {and} \bibinfo{person}{Aparna Taneja}.} \bibinfo{year}{2023}\natexlab{}.
\newblock \showarticletitle{Deployed SAHELI: Field Optimization of Intelligent
  RMAB for Maternal and Child Care}. In \bibinfo{booktitle}{\emph{Innovative
  Applications of Artificial Intelligence (IAAI)}}.
\newblock


\bibitem[Wood et~al\mbox{.}(2019)]%
        {wood2019taking}
\bibfield{author}{\bibinfo{person}{Christopher~S Wood},
  \bibinfo{person}{Michael~R Thomas}, \bibinfo{person}{Jobie Budd},
  \bibinfo{person}{Tivani~P Mashamba-Thompson}, \bibinfo{person}{Kobus Herbst},
  \bibinfo{person}{Deenan Pillay}, \bibinfo{person}{Rosanna~W Peeling},
  \bibinfo{person}{Anne~M Johnson}, \bibinfo{person}{Rachel~A McKendry}, {and}
  \bibinfo{person}{Molly~M Stevens}.} \bibinfo{year}{2019}\natexlab{}.
\newblock \showarticletitle{Taking connected mobile-health diagnostics of
  infectious diseases to the field}.
\newblock \bibinfo{journal}{\emph{Nature}} \bibinfo{volume}{566},
  \bibinfo{number}{7745} (\bibinfo{year}{2019}), \bibinfo{pages}{467--474}.
\newblock


\end{thebibliography}

\end{document}